\documentclass[journal]{IEEEtran}
\usepackage{amsmath}
\usepackage{multirow}
\usepackage[table,xcdraw]{xcolor}
\usepackage[normalem]{ulem}
\useunder{\uline}{\ul}{}
\usepackage{float}
\usepackage{cite}
\usepackage{url}
\usepackage{times} 
\usepackage[numbers,sort&compress]{natbib}

\usepackage{blindtext}
\usepackage{gensymb}
\usepackage{enumitem}
\usepackage{academicons}
\usepackage{hyperref}

\DeclareMathOperator*{\argmin}{arg\,min}

\ifCLASSINFOpdf
   \usepackage[pdftex]{graphicx}
\else
\fi
\begin{document}
%
\title{Multimodal Shared Autonomy for Social Navigation Assistance of Telepresence Robots}
%
%
%

\author{Kenechukwu C.~Mbanisi,~\IEEEmembership{Student Member,~IEEE,}
        Michael A.~Gennert,~\IEEEmembership{Senior Member,~IEEE}

\thanks{Kenechukwu C. Mbanisi and Michael A. Gennert are with the Robotics Engineering Department, Worcester Polytechnic Institute, Worcester, MA 01605, USA (email: {kcmbanisi, michaelg}@wpi.edu)}
}

%
%

\markboth{Journal of \LaTeX\ Class Files,~Vol.~14, No.~8, August~2015}%
{Mbanisi \MakeLowercase{\textit{et al.}}: Multimodal Shared Autonomy for Social Navigation Assistance of Telepresence Robots}
%



\maketitle

\begin{abstract}
Mobile telepresence robots (MTRs) have become increasingly popular in the expanding world of remote work, providing new avenues for people to actively participate in activities at a distance. However, humans operating MTRs often have difficulty navigating in densely populated environments due to limited situation awareness and narrow field-of-view, which reduces user acceptance and satisfaction. Shared autonomy in navigation has been studied primarily in static environments or in situations where only one pedestrian interacts with the robot. We present a multimodal shared autonomy approach, leveraging visual and haptic guidance, to provide navigation assistance for remote operators in densely-populated environments. It uses a modified form of reciprocal velocity obstacles for generating safe control inputs while taking social proxemics constraints into account. Two different visual guidance designs, as well as haptic force rendering, were proposed to convey safe control input. We conducted a user study to compare the merits and limitations of multimodal navigation assistance to haptic or visual assistance alone on a shared navigation task. The study involved 15 participants operating a virtual telepresence robot in a virtual hall with moving pedestrians, using the different assistance modalities. We evaluated navigation performance, transparency and cooperation, as well as user preferences. Our results showed that participants preferred multimodal assistance with a visual guidance trajectory over haptic or visual modalities alone, although it had no impact on navigation performance. Additionally, we found that visual guidance trajectories conveyed a higher degree of understanding and cooperation than equivalent haptic cues in a navigation task.
\end{abstract}


%
\IEEEpeerreviewmaketitle

\section{Introduction}
%
%
%
%
\IEEEPARstart{M}{obile} telepresence robots (MTRs) enable people to extend their presence to remote locations and provide new opportunities for remote participation. Recent years have seen an increase in adoption of these robots throughout various spheres of life. Today, MTRs make it possible for people to remotely participate in corporate meetings, academic conferences, college classes, elder care visits, and even medical appointments~\cite{Neustaedter2018, Khojasteh2019}. Advances in immersive technologies (e.g. Virtual Reality (VR), Head-mounted displays), as well as sensing and computing capabilities applied to telepresence robots continue to improve their ease of use in navigation and overall user experience~\cite{mimnaugh2021analysis}.

However, one of the challenges with MTRs is navigating around humans in cluttered environments without having a clear picture of the surroundings, due to their limited situational awareness and narrow field of view~\cite{Neustaedter2018}. Furthermore, it reduces their ability to pay attention to non-navigation-related activities, such as interacting with others or exploring the remote space.

One way to address this challenge is to provide fully autonomous features to handle the low-level navigation task. Considerable work has been done on autonomous socially-aware navigation algorithms, such as robust pedestrian motion prediction and tracking and dynamic collision avoidance~\cite{Kruse2013, mavrogiannis2021core}. However, these fully autonomous systems are yet to be fully reliable, especially in safety or time-critical scenarios, and may require human intervention in edge cases. Additionally, research has shown that telepresence robot operators may prefer to remain in the control loop for reasons such as a sense of agency, control and limited trust in autonomy~\cite{Khojasteh2019, Takayama2011}. The concept of shared autonomy, however, is a middle ground between full autonomy and manual control by combining the control inputs from the human operator and an autonomous agent to perform a shared navigation task, while retaining the final control authority with the operator as a safety precaution~\cite{Abbink2018}.


Studies have shown that haptic shared autonomy in assisted navigation improves driving performance~\cite{Petermeijer2015}, as well as situation awareness~\cite{Abbink2018}. Yet, these systems struggle with the issue of conflict between the human and the autonomous agent often resulting from a lack of alignment of shared intent or control strategy among cooperating agents~\cite{Itoh2016}. One way to mitigate this intent misalignment is by improving system transparency, i.e. how well the human operator can understand the intent and anticipate the actions of the autonomous agent~\cite{Muelling2018}. The shared haptic channel is a natural option for communicating intent as it leads to fast, reflexive responses. However, due to limited bandwidth, it may only be applicable in situations where the task is clear and unambiguous~\cite{de2007design}. As a result, researchers have favored visualization to augment haptic cues in intent communication. While the visual channel is slower in processing stimuli than the haptic channel, its trade-off is more bandwidth to convey richer, contextual information~\cite{Kuiper2016}. Thus, several works have proposed using visualization to communicate the intent of the autonomous agent in human-automation interaction. Augmented reality visualization has been applied for wheelchair navigation assistance~\cite{Zolotas2019}, assisted telemanipulation to communicate the agent's estimate of the desired goal~\cite{Brooks2020}. Recent studies have explored the benefits of combining both haptic and visual information in a multi-modal fashion~\cite{Vreugdenhil2019, van2020visual}.



In this paper, we present a multimodal shared autonomy approach in the context of social navigation assistance. Our main contributions are as follows: 

(i) We present a modified reciprocal velocity obstacle (RVO) approach for social navigation assistance that generates safe control signals for the human operator by considering social proxemics constraints~\cite{hall1966hidden}. This extends the state-of-the-art in assisted navigation of telepresence robots from static environments to dynamic, human-populated environments. 

(ii) We implemented and evaluated different feedback modalities (haptics, visual, and a combination of both) in a user study to assess the effects on navigation performance, system transparency, cooperation and user preference. 

This paper extends our work-in-progress paper~\cite{socnavassist21}, which only described the concept underlying the proposed approach.



\section{Related Work}\label{sec2}

\subsection{Overview of Shared Autonomy for Assisted Navigation}

Among the different types of autonomy, shared autonomy sits in the middle between manual control and full autonomy~\cite{Beer2014}. Abbink et al.~\cite{Abbink2018} described shared autonomy (also known as shared control) as a scenario where ``humans and robots are interacting congruently in a perception-action cycle to perform a dynamic task that either the human or the robot could execute individually under ideal circumstances." With shared autonomy, autonomous agents can contribute to the execution of tasks, resulting in better performance and reduced operator workloads in passenger vehicle driving~\cite{Petermeijer2015}. 
In cluttered indoor environments, semi-autonomous features can also reduce operator burden and improve driving performance. They are used both in co-located systems such as smart wheelchair systems~\cite{Zolotas2019} and in remote systems such as mobile telepresence robots~\cite{Takayama2011}. The majority of existing work has concentrated on static environments, with little attention paid to assistance in dynamic, pedestrian-rich environments, while studies have found that navigating a vehicle in human-crowded spaces is difficult~\cite{Neustaedter2018}. 

Recent years have seen considerable interest in the topic of autonomous socially-aware robot navigation (for a review, see~\cite{mavrogiannis2021core}). Several socially-aware planning approaches have been proposed in the literature and the prominent categories include sampling-based (e.g., RiskRRT~\cite{rios2011understanding}), model-based (e.g., social force model (SFM)~\cite{helbing1995social}, velocity-obstacles (VO)~\cite{VanBerg2008}) and learning-based methods (e.g., SA-CADRL~\cite{Chen2017}). However, there has been limited application of these methods to driving assistance, where the human is in the loop. In \cite{Narayanan2016}, Narayanan et al formulated a semi-autonomous navigation framework for smart wheelchair assistance using RiskRRT. Kretzschmar et al. \cite{Kretzschmar2016} employed an inverse reinforcement learning method to learn the social constraints inherent in navigation from human demonstration, and then used this trained policy as an assistive-agent on a wheelchair. We extend these systems to exploring the social interaction problem with multiple pedestrians, which is more realistic and general. Moreover, we consider a remote teleoperation scenario in which a robot is physically separated from the operator, causing reduced situational awareness and narrow field of view. Therefore, navigation assistance plays an increasingly important role in social navigation.

\subsection{Feedback Modalities in Shared Autonomy}



Studies have investigated various feedback modalities to enable the autonomous agent to communicate with a human, with the most prominent options being haptics and vision~\cite{hoc2009cooperation}.
Haptic feedback has been researched extensively in the automotive domain and has found broad application in assisted driving tasks (e.g., lane-keeping assist)~\cite{Abbink2018}. Haptic cues are generally provided in two modes: vibrotactile feedback and haptic force feedback~\cite{losey2018review}. Haptic shared control provides haptic torques or forces via haptic-enabled control interface (e.g., motorized steering wheel) to guide an operator driving towards a reference path~\cite{Abbink} or away from forbidden regions (e.g., obstacles)~\cite{Zhang2020}. Studies show that haptic feedback is effective in communicating instantaneous motion corrections because it elicits fast reflex responses~\cite{de2007design}. However, it is limited in the ability to convey contextual information such as the future actions or intent of the autonomous agent~\cite{Vreugdenhil2019}.

On the other hand, the visual channel is able to convey more contextual information, but less effective in time-critical tasks due to shower response times~\cite{Kuiper2016}. Visual feedback has been applied in advanced driver assistance systems (ADAS) for warning support (e.g., lane departure warning) as well as in action suggestion (e.g., recommended route display)~\cite{hoc2009cooperation}. Recent studies have also enabled visual guidance in non-screen applications such as wheelchair systems using head-mounted displays (HMDs)~\cite{Zolotas2019}.

Research has considered multimodal feedback combining visual and haptic modalities to improve intent understanding. Vreugdenhil et al.~\cite{Vreugdenhil2019} found that multimodal assistance improved user acceptance over haptic feedback alone. Berg~\cite{van2020visual} showed that compensating haptic forces with a visualization increased task performance in an industrial application. In a UAV collision avoidance task experiment, Ho et al.~\cite{ho2018increasing} evaluated two visualization designs combined with haptic feedback. They reported that participants preferred the visual design that showed a 1:1 representation of haptic forces. However, multimodal feedback may not always improve performance and may be dependent on the complexity of the task~\cite{losey2018review}. Our study examines whether multimodal feedback has a positive impact on a shared navigation task in a dynamic environment.

\section{System Design}\label{sec:system-design}

Our socially-aware navigation assistance approach incorporates three key elements: (1) socially-aware collision avoidance via reciprocal velocity obstacles (SA-RVO), (2) guidance using haptic forces, and (3) guidance using visual cues.

\subsection{SA-RVO: Socially-aware collision avoidance via reciprocal velocity obstacles}\label{subsec:sa-rvo}

To enable effective navigation assistance in dynamic environments, it is crucial to generate safe control signals that can guide human operators towards socially acceptable and collision-free movements. We base our approach on the reciprocal velocity obstacle (RVO) method~\cite{VanBerg2008}. RVO and its variants have been successfully implemented in multi-robot systems and autonomous socially-aware navigation~\cite{Truong2017}. RVO is an extension of the classic velocity obstacle method of dynamic collision avoidance, which guarantees collision- and oscillation-free motions in the presence of moving obstacles (such as other robot agents). This is accomplished by planning motions in the 2D velocity space of the robot and surrounding static or dynamic obstacles. Specifically, the robot $A$ measures the relative velocities of other agents (i.e. pedestrian $B$) within its vicinity and constructs a collision cone ($RVO_{A|B}$) which is a region in the velocity space that would lead to a collision with the other agent~\cite{VanBerg2008} (Fig.~\ref{fig:socnavassist-rvo}).  The apex position of the collision cone, $RVO_{A|B}$, is translated to $(1-\alpha) V_{A} + \alpha V_B$ in velocity space, where $\alpha \in [0,1]$ refers to the share of collision avoidance responsibility that the robot agent takes to avoid the pedestrian~\cite{VanBerg2008}.

In the original RVO formulation, the optimal velocity is calculated as the candidate velocity within the space of collision-free velocities (i.e. $v \notin RVO_{A|B}$) that minimizes the distance to the maximum allowable velocity to the goal as follows:

\begin{figure}[t]
        \centering
        \includegraphics[width=0.98\columnwidth]{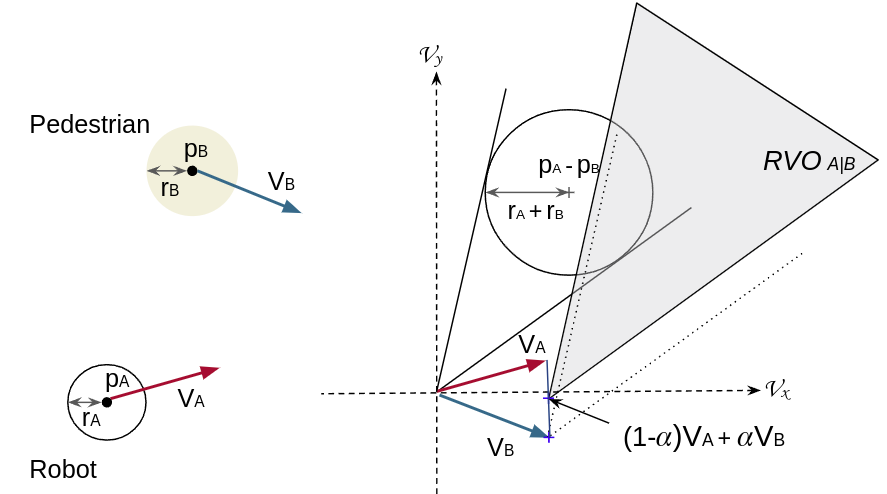}
        \caption{Schematic describing the reciprocal velocity obstacles (RVO) formulation. $RVO_{A|B}$ is the collision cone defined as the region of velocities for robot A that would result in a collision with pedestrian $B$.}
        \label{fig:socnavassist-rvo}        
\end{figure}

\begin{equation}
    v_A^{optimal}(t) = \argmin_{v \notin RVO_{A|B}} \| v(t) - v^{goal}(t) \|^2
    \label{rvo-equ}
\end{equation}

\noindent
where $v^{goal}(t)$ is the maximum velocity to the goal location~\cite{VanBerg2008}.

We modify the search for optimum velocity by taking into account additional factors for navigation assistance, resulting in a weighted sum of objective functions:

\begin{equation}
    v_A^{optimal}(t) = \argmin_{v \notin RVO_{A|B}} \sum_{i=1}^{n} w_i G_i
\end{equation}

\vspace{2mm}

\noindent
where $w_i$ and $G_i$ are the $i$th weight and objective function respectively. In this work, we consider three objective functions ($n=3$), with $G_1 = \| v(t) - v_A^{pref}(t) \|^2$ 
where $v_A^{pref}$ represents the current input velocity command from the human operator. This objective looks for a velocity that is as close as possible to the operator's commanded velocity, preserving the original control intent of the operator.
The second objective, $ G_2 = \| v(t) - v_A^{optimal}(t-1) \|^2$, controls for change in consecutive optimal velocities to enable smooth guidance cues to the operator. Finally, the third objective compensates for goal-directed motion as in (\ref{rvo-equ}), $G_3 = \| v(t) - v^{goal}(t) \|^2$. This requires that the autonomous system is aware of the operator's goal either explicitly or by inference~\cite{jain2019probabilistic}. For this study, we assume that the operator's goal location is known.

The RVO formulation only accounts for disc-shaped objects, therefore, we perform a static obstacle avoidance step as a preprocessing step to the RVO, in order to account for static obstacles in the environment such as walls and tables. Specifically, using velocity sampling, we filter out velocities that would lead to collision over a time horizon based on a pre-defined map of the environment. The resulting samples of `statically' safe velocities is then fed into the SA-RVO to account for dynamic obstacles.

In addition, our proposed approach follows Truong et al.~\cite{Truong2017} by explicitly accounting for social proxemics constraints. Based on the proxemics theory proposed by Hall~\cite{hall1966hidden}, an individual's interpersonal space can be divided into concentric circles, with the radius of the circles representing different levels of intimacy and comfort: intimate space ($<$0.45m), personal space (0.45-1.2m) and then social space ($>$1.2m). Hence, SA-RVO defines the collision cone for each pedestrian based on the radius of their personal space, thus taking into account their social constraints.

Furthermore, in addition to the proxemics constraints imposed on single pedestrians, we look at human group interactions and their corresponding social constraints. Human groups are detected and tracked based on the multi-model multiple hypothesis tracker (MHT) approach proposed by Linder and Arras~\cite{linder2014multi}. The open-source ROS implementation\footnote{\url{https://github.com/spencer-project/spencer_people_tracking}} was used in this study.

\subsection{Guidance using Haptic forces}\label{subsec:haptic}

The haptic guidance mode generates appropriate haptic feedback forces on the operator control interface (i.e. force-feedback haptic joystick) used by the operator to control the robot. The generated haptic forces, $F(t)$, guide the operator’s control input towards alignment with the optimal velocity command, $v_A^{optimal}(t)$, computed by SA-RVO.  The haptic forces are computed as proportional to the error between $v_A^{optimal}(t)$ and $v_A^{pref}$, which is the current velocity command from the human operator:

\begin{equation}
    F(t) = K_p(v_A^{optimal}(t) - v_A^{pref}(t))
\end{equation}

\vspace{3mm}

\noindent
where $K_p$ is the fixed haptic gain. This value regulates the level of haptic authority assigned to the autonomous agent. With a low haptic gain, the operator is able to override low haptic guidance forces easily, thereby retaining control. A high haptic gain, however, may make it difficult for the operator to counteract the haptic forces being generated, leaving final control to the autonomous agent~\cite{Abbink}.

\begin{figure}[t]
        \centering
        \includegraphics[width=0.95\columnwidth]{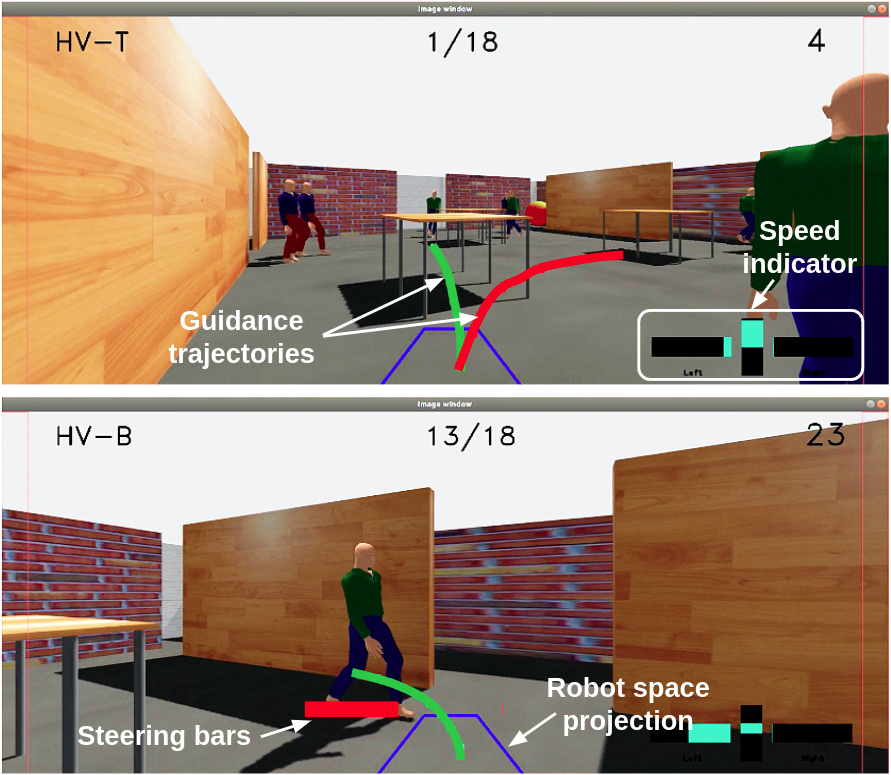}
        \caption{Illustration of the visual guidance cues. \textbf{Top}: visual guidance trajectory. \textbf{Bottom}: visual steering bars. In both designs, the green trajectory trace presents a feedback on the dynamics of the robot, while the guidance cues are in red. A projection of the robot's occupied space is in blue, and a speed indicator display is in aqua on the bottom right.}
        \label{fig:visual}        
\end{figure}

\subsection{Guidance using Visual cues}\label{subsec:visual}

Visual feedback in mobile telepresence robots typically serves two purposes. It provides the operator with information about the remote environment in which the robot operates, and it also enables audio-visual social interaction with the remote individuals. However, existing studies have shown that augmented visualizations on the visual feedback display can be used to provide guidance cues~\cite{Kuiper2016, Zolotas2019} and communicate the intent of the autonomous system to the human operator~\cite{brooks2020visualization, evans2015investigating}.

In this study, we examine two visual cue designs. First, a visual trajectory trace is presented as an overlay to communicate the predicted future states of robot based on the instantaneous control input from the shared haptic interface as in~\cite{Vreugdenhil2019}. Predicted trajectories are computed using a simple kinematics model of the differential drive robot, which propagates the current state by the control input. This helps the operator form an improved mental model of the robot dynamics. 

Secondly, we seek to provide visual cues on the visual display to guide the operator towards the optimal control input computed by SA-RVO and consistent with the haptic guidance mode. To achieve this, we consider two visualization designs:

\begin{enumerate}
    \item \textit{Visual guidance trajectory}: Inspired by~\cite{Vreugdenhil2019}, we also present the visual guidance cue as a visual trajectory trace as above. In contrast to~\cite{Vreugdenhil2019} however, the predicted robot future states are computed based on the optimal control command computed by SA-RVO rather than the operator's control input. The result is a suggested path visualization that shows contextual information about the direction the autonomous agent suggests the operator to go. Note that we display the trajectory only when the difference between the operator's control input and the optimal control input exceeds a threshold, in order to avoid display clutter.
    
    \item \textit{Visual steering bars}: Instead of showing a suggested path, this approach presents steering magnitude bars on the left and right of the visual display (Fig.~\ref{fig:visual}). They indicate how much the robot needs to be steered in order to align with SA-RVO's optimal control inputs. In this design, instantaneous steering corrections consistent with the haptic cues are communicated to the operator over the visual display. This provides the operator with limited information about the suggested future states and may be viewed as a visual representation of the haptic cues. 
    
\end{enumerate}


\section{User Study Design}\label{sec:user-study-design}

We conducted a user study with a within-subject repeated-measures design to compare the effects of multimodal navigation assistance with single-modality (haptic and visual) and no assistance cases, as a baseline, in terms of navigation performance, interface transparency and user preference in a social navigation task. The study design and procedure was duly approved by Worcester Polytechnic Institute's Institutional Review Board (IRB). 

Our study hypotheses are as follows: First, we hypothesize that multimodal assistance will outperform both haptic and visual modalities alone in regards to navigation performance, system transparency and cooperation, and user preference (\textbf{H1}). Additionally, we anticipate that all guidance modes will outperform manual control in navigation performance and user preference when considered as a baseline condition (\textbf{H2}). Furthermore, we hypothesize that visual modality will lead to greater system transparency and cooperation than haptic modality (\textbf{H3}).




\begin{figure*}[t]
        \centering
        \includegraphics[width=1.5\columnwidth]{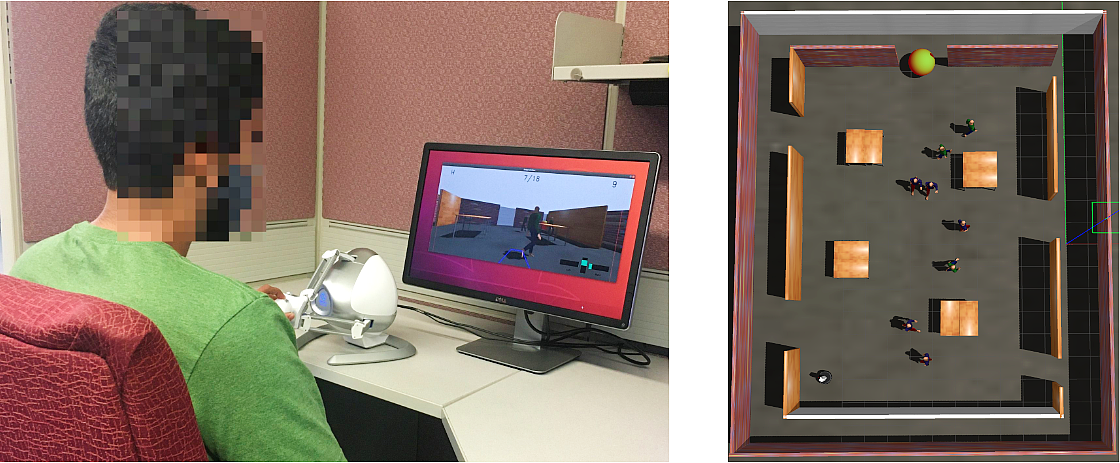}
        \caption{ Simulated experiment setup. \textit{Left:} Each participant controls a virtual wheeled robot using a haptic joystick with visual feedback from the forward-facing camera on the robot \textit{Right:} Snapshot of the simulated social navigation task.}
        \label{fig:socnavassist-exp}        
\end{figure*}

\subsection{Experimental Setup}\label{subsec:setup}

The user study was designed as a virtual social navigation task. Fifteen (15) participants (7 males and 8 females; age: 25.3 $\pm$ 4.1) were recruited for the user study from the university student community.
Twelve participants (80\%) reported to have no prior experience with a haptic device. 

Participants were asked to control a virtual mobile telepresence robot in a dynamic, human-populated virtual environment to navigate from start position in a hall to specified goal location in a safe and socially acceptable manner. The virtual mobile robot and the populated environment were simulated using Gazebo Simulator.  The virtual mobile robot is based on the Freight mobile base (Fetch Robotics) which is a differential drive wheeled UGV with support castor wheels~\cite{wise2016fetch}.
The differential drive kinematics and dynamic model follow the nonholonomic non-slip constraints and are based on~\cite{fukao2000adaptive}. We used the \textit{diff-drive controller}\footnote{\url{http://wiki.ros.org/diff_drive_controller}} in ROS to control the robot with linear and angular velocities as the 2D control input. The control input and haptic guidance forces were applied through a commercially available force-feedback joystick (Novint Falcon, Novint Technologies). Given the 2D control input, only two of the 3-DOFs on the haptic joystick were mapped to the robot controller using a position-velocity mapping.

Participants received visual feedback from a forward-facing camera mounted on the virtual mobile robot through a 24-inch computer monitor (see Fig.~\ref{fig:socnavassist-exp}). The visual guidance methods were presented as a visual overlay on the camera display using OpenCV. The virtual environment was modeled after a conference hall with tables and space for walking (see Fig.~\ref{fig:socnavassist-exp}). Two different hall layouts were adopted and randomly applied based on table positions to prevent participants from memorizing the hall layout across trials. We modeled the virtual pedestrian motion using a Gazebo actor plugin\footnote{\url{https://github.com/robotics-upo/gazebo_sfm_plugin}}\footnote{\url{https://github.com/robotics-upo/lightsfm}} based on the social force model~\cite{helbing1995social} to enable reactive navigation behavior to obstacles and the robot. Three pedestrian motion configurations were implemented in the study: (a) approach, (b) crossing, (c) random, to encompass a wide variety of real-life crowd navigation scenarios.


\subsection{Evaluated Conditions}\label{subsec:conditions}

In the study, participants drove the virtual mobile robot in six control conditions based on the presence and form of navigation assistance provided. They are as followed:

\begin{enumerate}
    \item \textbf{Manual control (MC)}: This is the baseline condition with no navigation assistance.
    \item Single modal assistance:
    \begin{enumerate}
        \item \textbf{Haptic guidance (H)}: Navigation assistance is provided in the form of haptic guidance forces on the haptic interface based on  Section~\ref{subsec:haptic}. 
        \item \textbf{Visual guidance trajectory (V-T)}: A suggested path visualization is presented along with the predicted robot path as described in  Section~\ref{subsec:visual}. 
        \item \textbf{Visual steering bars (V-B)}: Steering correction cues are presented along with the predicted robot path as described in  Section~\ref{subsec:visual}. 
    \end{enumerate}
    \item Multimodal assistance:
    \begin{enumerate}
        \item \textbf{Haptic + Visual trajectory (HV-T)}: Combines the haptic guidance cues with the visual trajectory cues.
        \item \textbf{Haptic + Visual bars (HV-B)}: Combines the haptic guidance cues with the visual steering bars.
    \end{enumerate}
\end{enumerate}


\subsection{Experimental Procedure}\label{subsec:procedure}


The study commenced with an introduction to the study procedure and how to control the virtual mobile robot. The study comprised three phases: familiarization, learning, and testing phases.

\textit{Familiarization Phase:} Here, participants were allowed to drive the virtual mobile robot in an open space (with no obstacles or people) for up to 5-7 minutes to get familiar with the robot controls. Participants held the haptic control interface using their preferred hand. After participants report confidence in their control of the robot, they are moved to the next phase.

\textit{Learning Phase:} In this phase, the participants completed 12 trials of the navigation task. In each trial, they were told to control the robot from its start location in the virtual environment to a specified goal location within the hall while driving in a socially-acceptable manner without colliding with moving pedestrians and obstacles.
A trial starts when the participant started the timer by pressing a button on the haptic interface and ended once the robot reaches within a threshold distance of the goal location. Each trial was randomly assigned an evaluated condition, resulting in participants experiencing all six conditions at least twice. This allowed participants to build familiarity with the assisted control conditions before the main testing phase. A break time option was provided to avoid fatigue during the study.

\textit{Testing Phase:} In this phase, the participants completed a block of three trials for each control condition, resulting in six blocks and 18 trials in total. Each trial block comprised the three pedestrian configurations described above. The order of the conditions were randomized and counterbalanced to reduce learning effect, recency bias and fatigue. After each block, participants completed a post-block questionnaire to provide subjective responses to their experience. At the completion of all six trial blocks, participants completed a post-study questionnaire where they provides subjective ranking of the six control conditions according to several criteria.  Section~\ref{subsec:measures}  provides more details on the two questionnaires administered. The entire user study took approximately 1.5 hours to complete, depending on the participants.






\subsection{Measures}\label{subsec:measures}

\subsubsection{Objective Measures}
Five objective measures were computed to evaluate navigation performance, and the level of cooperation between the human and the navigation assistance.

\begin{enumerate}
    \item \textit{Number of intimate and personal intrusions}: The number of instances of intimate and personal space intrusions per trial. An intrusion occurs when the robot-to-pedestrian clearance is below 0.45m and 1.2m for intimate and personal spaces respectively~\cite{rios2015proxemics}.

    
    \item  \textit{Path length (metres)}: Total distance travelled by the robot in the trial.
    
    \item \textit{Trial time (secs)}: Total time taken to complete the trial.
    
    \item \textit{Mean disagreement}: The mean of the norm of the difference between the human operator's velocity input and the optimal velocity input from the navigation assistant~\cite{Zhang2020}.
\end{enumerate}

\subsubsection{Subjective Measures}

\paragraph{Post-block Questionnaire} At the end of each trial block with each control condition, participants completed the following questionnaire using a 7-point likert scale to evaluate the condition:
\begin{enumerate}
    \item \textit{System helpfuless and ease of use}: (1 - Strongly disagree, 7 - Strongly agree)
    \begin{itemize}
        \item \textit{Helpfulness}: ``This interface condition was very helpful for completing the navigation task."
        \item \textit{Ease of use}: ``I found this interface condition to be easy to use for navigating the robot."
    \end{itemize}
    \item \textit{System transparency}: (1 - Understood nothing at all, 7 - Understood extremely well)
    \begin{itemize}
        \item \textit{Intent understanding}: ``To what extent did you understand the intentions of the navigation assistance?" 
        \item \textit{Force anticipation}: ``To what extent could you anticipate the haptic forces provided by the navigation assistance?"
    \end{itemize}
    \item \textit{Level of cooperation}: ``To what extent did you cooperate or agree with the navigation assistance in the task?" (1 - Never agreed, 7 - Always agreed)
    \item \textit{Sense of control}: ``To what extent did you feel you had control over the robot?" (1 - No control, 7 - Complete control)
\end{enumerate}

\paragraph{Post-study Questionnaire and Interview} After all trial blocks, participants completed the following questionnaire to rank each of the control conditions.
\begin{enumerate}
    \item \textit{Skill confidence level}: ``How confident are you that you can complete the task successfully without navigation assistance?" (1 - Not confident at all, 7 - Extremely confident)
    \item Rank the six conditions in terms of the \textit{most helpful} in completing the tasks successfully.
    \item Rank the five assisted conditions in terms of the \textit{most intent understanding} of the navigation assistance.
    \item Rank the five assisted conditions in terms of the \textit{most cooperation (or agreement)} with the navigation assistance.
    \item Rank the six conditions in terms of the \textit{most preferred} (considering all factors) for completing the task.
\end{enumerate}
At the end of the study, we conducted a short interview for each participant to understand the reasons for how they ranked the conditions and other follow-up questions.

\begin{table*}[]
\small
\centering
\caption{Means (M), standard deviations (SD) and statistical analysis results for the objective and subjective measures across control conditions. The symbol $^-$ denotes the measure is of negative scale.}
\begin{tabular}{cccccccclc}
\hline
                                                & \multicolumn{1}{l}{}      & \multicolumn{6}{c}{\textbf{Conditions}}                                                                                                                                                 &  & \multicolumn{1}{l}{}                                         \\ 
\multirow{-2}{*}{\textbf{Measures}}             &                           & MC                           & H                            & V-T                          & V-B                          & HV-T                         & HV-B                         &  & \multicolumn{1}{l}{\multirow{-2}{*}{\textbf{Friedman Test}}} \\ \hline \hline
                                                & M                         & 0.40                         & 0.51                         & 0.80                         & 0.49                         & 0.60                         & 0.53                         &  & \textit{p = 0.240}                                           \\
\multirow{-2}{*}{Intimate intrusions$^-$} & {\color[HTML]{656565} SD} & {\color[HTML]{656565} 0.315} & {\color[HTML]{656565} 0.280} & {\color[HTML]{656565} 0.450} & {\color[HTML]{656565} 0.354} & {\color[HTML]{656565} 0.422} & {\color[HTML]{656565} 0.451} &  & {\color[HTML]{656565} \textit{$\chi^2$(5) = 6.741}}                 \\ \hline
                                                & M                         & 2.38                         & 3.134                         & 3.089                         & 2.711                         & {\textbf{2.245}}            & 3.267                         &  & \textit{\textbf{p = 0.067*}}                        \\
\multirow{-2}{*}{Personal intrusions$^-$} & {\color[HTML]{656565} SD} & {\color[HTML]{656565} 1.033} & {\color[HTML]{656565} 1.464} & {\color[HTML]{656565} 1.890} & {\color[HTML]{656565} 1.320} & {\color[HTML]{656565} 1.642} & {\color[HTML]{656565} 1.733} &  & {\color[HTML]{656565} \textit{$\chi^2$(5) = 10.279}}                 \\ \hline

                                                & M                         & 17.99                         & 19.50                         & 19.05                         & 17.95                        &  18.02            & 19.07                         &  & \textit{p = 0.561}                        \\
\multirow{-2}{*}{Path length (m)} & {\color[HTML]{656565} SD} & {\color[HTML]{656565} 1.92} & {\color[HTML]{656565} 2.55} & {\color[HTML]{656565} 2.39} & {\color[HTML]{656565} 2.18} & {\color[HTML]{656565} 2.24} & {\color[HTML]{656565} 3.39} &  & {\color[HTML]{656565} \textit{$\chi^2$(5) = 3.914}}                 \\ \hline

                                                & M                         & 41.57                         & 59.79                         & 44.21                         & 53.90                        &  49.68            & 50.80                         &  & \textit{p = 0.199}                        \\
\multirow{-2}{*}{Trial time (s)} & {\color[HTML]{656565} SD} & {\color[HTML]{656565} 14.73} & {\color[HTML]{656565} 34.16} & {\color[HTML]{656565} 12.88} & {\color[HTML]{656565} 36.35} & {\color[HTML]{656565} 26.69} & {\color[HTML]{656565} 26.22} &  & {\color[HTML]{656565} \textit{$\chi^2$(5) = 7.29}}                 \\ \hline



                                                & M                         & -                            & 0.37                         & 0.28                         & 0.30                         & 0.30                         & 0.28                         &  & \textit{p = 0.735}                                           \\
\multirow{-2}{*}{Mean disagreement$^-$}             & {\color[HTML]{656565} SD} & {\color[HTML]{656565} -}     & {\color[HTML]{656565} 0.186} & {\color[HTML]{656565} 0.121} & {\color[HTML]{656565} 0.152} & {\color[HTML]{656565} 0.183} & {\color[HTML]{656565} 0.112} &  & {\color[HTML]{656565} \textit{$\chi^2$(4) = 2.000}}                 \\ \hline
                                                & M                         & 5.33                         & 4.27                         & {\textbf{5.73}}                   & 5.20                         & 5.40                         & 5.40                         &  & \textit{\textbf{p = 0.026}}                                  \\
\multirow{-2}{*}{Interface helpfulness}         & {\color[HTML]{656565} SD} & {\color[HTML]{656565} 1.633} & {\color[HTML]{656565} 1.831} & {\color[HTML]{656565} 0.961} & {\color[HTML]{656565} 1.207} & {\color[HTML]{656565} 1.595} & {\color[HTML]{656565} 1.352} &  & {\color[HTML]{656565} \textit{$\chi^2$(5) = 12.68}}                 \\ \hline
                                                & M                         & 5.80                         & 4.27                         & {\textbf{6.07}}                   & 5.67                         & 5.40                         & 5.40                         &  & \textit{\textbf{p = 0.012}}                                  \\
\multirow{-2}{*}{Interface ease of use}         & {\color[HTML]{656565} SD} & {\color[HTML]{656565} 1.639} & {\color[HTML]{656565} 1.831} & {\color[HTML]{656565} 1.100} & {\color[HTML]{656565} 0.976} & {\color[HTML]{656565} 1.765} & {\color[HTML]{656565} 1.639} &  & {\color[HTML]{656565} \textit{$\chi^2$(5) = 14.51}}                 \\ \hline
Intent understanding                            & M                         & -                            & 3.73                         & {\textbf{5.80}}                   & 5.40                         & 5.33                         & 5.47                         &  & \textit{\textbf{p = 0.008}}                                  \\
(Transparency)                                  & {\color[HTML]{656565} SD} & {\color[HTML]{656565} -}     & {\color[HTML]{656565} 2.120} & {\color[HTML]{656565} 1.146} & {\color[HTML]{656565} 1.404} & {\color[HTML]{656565} 1.799} & {\color[HTML]{656565} 1.506} &  & {\color[HTML]{656565} \textit{$\chi^2$(4) = 13.72}}                 \\ \hline
Force Anticipation                              & M                         & -                            & 3.20                         & -                            & -                            & 3.93                         & {\textbf{4.47}}                   &  & \textit{\textbf{p = 0.049}}                                   \\
(Transparency)                                  & {\color[HTML]{656565} SD} & {\color[HTML]{656565} -}     & {\color[HTML]{656565} 1.935} & {\color[HTML]{656565} -}     & {\color[HTML]{656565} -}     & {\color[HTML]{656565} 1.710} & {\color[HTML]{656565} 1.922} &  & {\color[HTML]{656565} \textit{$\chi^2$(2) = 6.00}}                  \\ \hline
                                                & M                         & -                            & 3.67                         & {\textbf{5.267}}                  & 4.93                         & 4.93                         & 4.67                         &  & \textit{\textbf{p = 0.019}}                                  \\
\multirow{-2}{*}{Cooperation}                   & {\color[HTML]{656565} SD} & {\color[HTML]{656565} -}     & {\color[HTML]{656565} 1.345} & {\color[HTML]{656565} 0.961} & {\color[HTML]{656565} 1.280} & {\color[HTML]{656565} 1.580} & {\color[HTML]{656565} 1.345} &  & {\color[HTML]{656565} \textit{$\chi^2$(4) = 11.76}}                 \\ \hline
                                                & M                         & {\textbf{6.47}}                   & 4.80                         & 6.33                         & 6.27                         & 5.47                         & 5.40                         &  & \textit{\textbf{p = 0.000}}                                  \\
\multirow{-2}{*}{Sense of control}              & {\color[HTML]{656565} SD} & {\color[HTML]{656565} 0.915} & {\color[HTML]{656565} 1.474} & {\color[HTML]{656565} 0.724} & {\color[HTML]{656565} 0.704} & {\color[HTML]{656565} 1.125} & {\color[HTML]{656565} 1.29}  &  & {\color[HTML]{656565} \textit{$\chi^2$(5) = 27.89}}                 \\ \hline
\end{tabular}
\label{stats-results}
\end{table*}

\section{Results}\label{sec:results}


~
All statistical analyses were performed in R~\cite{rsoft}. Main effects across evaluated conditions was analyzed using the Friedman test (a non-parametric test for differences between multiple groups) and significance levels were estimated at $p < 0.05$ for statistical significance and $p < 0.1$ for marginal statistical significance. Post-hoc tests were performed using the Wilcoxon sign-test (a non-parametric test for differences between two dependent samples) with Bonferroni correction for multiple comparisons.~

A summary of statistical results is shown in Table~\ref{stats-results}. We scored the subjective rankings using the Borda count. The order of ranking across various criteria is presented in Table~\ref{rank-results}.

\begin{table*}[]
\small
\centering
\caption{Subjective ranking and calculated scores (in parentheses) of control conditions based on Borda count method}
\begin{tabular}{ccccccc}
\hline
\textbf{Borda Count Ranking} & \textbf{1st} & \textbf{2nd} & \textbf{3rd} & \textbf{4th} & \textbf{5th} & \textbf{6th} \\ \hline \hline
Helpfulness          & HV-T {\color[HTML]{656565} (73)}   & V-T {\color[HTML]{656565} (63)}         & HV-B {\color[HTML]{656565} (58)}        & V-B {\color[HTML]{656565} (51)}         & MC {\color[HTML]{656565} (49)}          & H  {\color[HTML]{656565} (21)}          \\

Intent Understanding & HV-T {\color[HTML]{656565} (63)}   & V-T {\color[HTML]{656565} (58)}         & HV-B {\color[HTML]{656565} (45)}        & V-B {\color[HTML]{656565} (40)}         &  H  {\color[HTML]{656565} (19)}  &     -     \\         

Cooperation          &  HV-T {\color[HTML]{656565} (57)}   & V-T {\color[HTML]{656565} (55)}         & V-B {\color[HTML]{656565} (48)}        & HV-B {\color[HTML]{656565} (42)}         &  H  {\color[HTML]{656565} (23)}  &     -     \\ 

Overall Preference   &  HV-T {\color[HTML]{656565} (68)}   & V-T {\color[HTML]{656565} (66)}         & V-B {\color[HTML]{656565} (55)}        & HV-B {\color[HTML]{656565} (54)}         & MC {\color[HTML]{656565} (50)}          & H  {\color[HTML]{656565} (22)}

\\ \hline
\end{tabular}
\label{rank-results}
\end{table*}

\begin{figure}[]
    \centering
    \includegraphics[width=0.80\columnwidth]{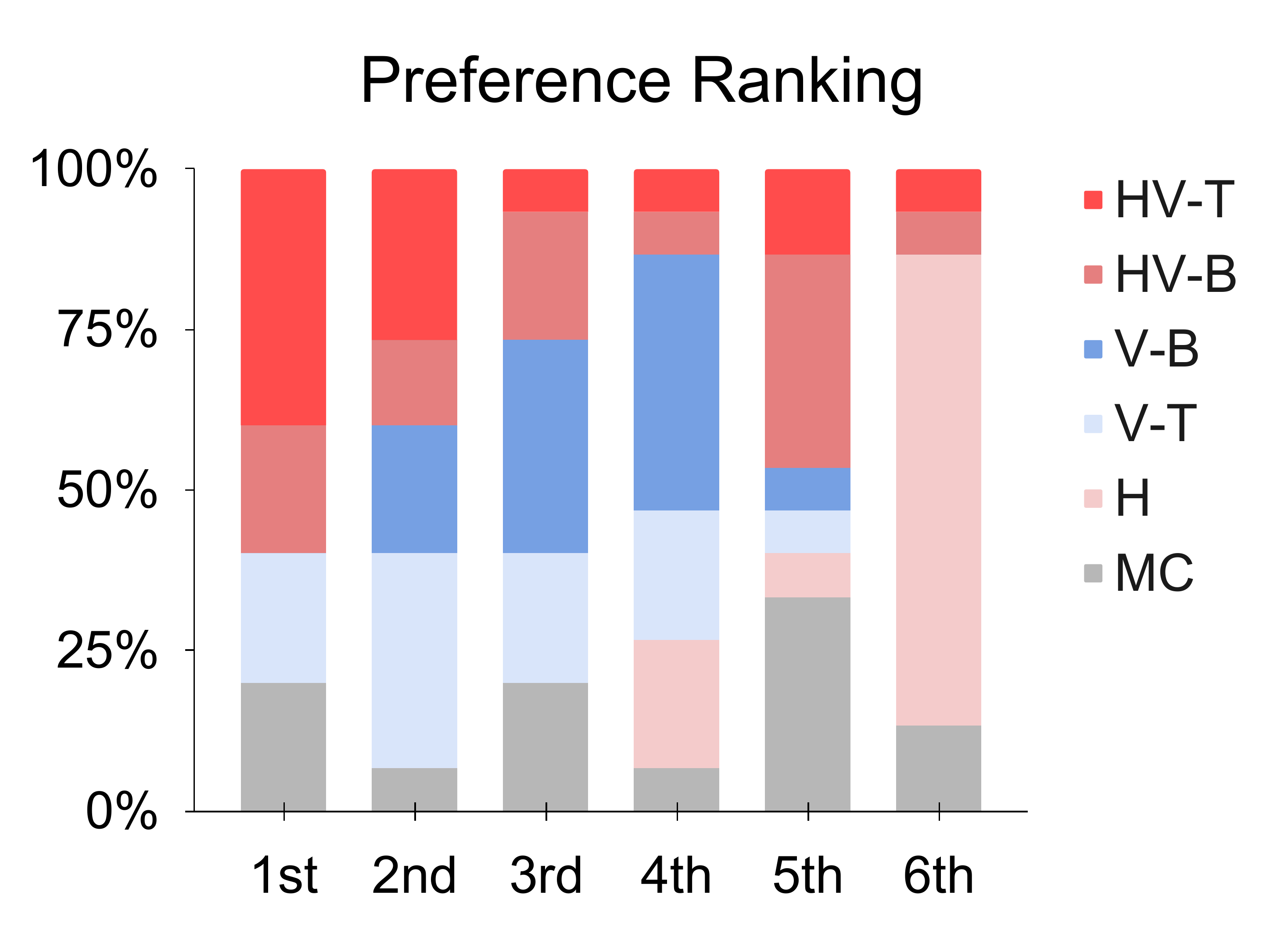}
    \caption{Distribution of overall preference ranking across control conditions.}
    \label{fig:preference}
\end{figure}

\begin{enumerate}
    \item  \textit{Navigation performance}: In evaluating the impact of navigation assistance on navigation performance (safety and efficiency), we found that control condition had no significant effect on the number of intimate intrusions, $(p = 0.357)$, the path length, $(p = 0.561)$, or trial time, $(p = 0.199)$, but had a marginally significant effect on the number of personal intrusions, $(p = 0.067)$. Furthermore, no significant effects were found in pairwise comparison between control conditions. It is noteworthy to mention that both safety and efficiency measures had very high variability across conditions, further reflecting a lack of substantial effect on both safety and navigation efficiency.
    
    \item \textit{System helpfulness and ease of use}: The subjective ratings of helpfulness and ease of use were compared across all six control conditions. We found statistical significance for both helpfulness, $\chi^2(5) = 12.68, p = 0.026$, and ease of use, $\chi^2(5) = 14.51, p = 0.012$ (Table~\ref{stats-results}). However, a post-hoc pairwise comparison found no statistical significance between the conditions. V-T has the highest mean rating for both helpfulness ($M = 5.73$) and ease of use ($M = 6.07$) among all the conditions. In the subjective ranking, HV-T was ranked most helpful 8 times (53\%), more than every other condition, leading to the highest score on the borda ranked-order count (see Table~\ref{rank-results}). Additionally, visual trajectory-based conditions (HV-T and V-T) were ranked most helpful a combined 11 times (73\%). 
    
    \item \textit{System transparency}: Our analysis on the transparency measures revealed statistical significance in intent understanding, $\chi^2(4) = 13.72, p = 0.008$, across the five assisted control conditions (H, V-T, V-B, HV-T, HV-B) and force anticipation, $\chi^2(2) = 6.0, p = 0.049$, across the haptic-based control conditions (H, HV-T, HV-B). Further post-hoc analysis found marginal difference between V-T and H ($p = 0.072$) for intent understanding and between HV-B and H ($p = 0.051$) for force anticipation. Other pairwise comparisons were not statistically significant. In the subjective ranking, the haptic (H) condition was ranked as hardest to understand 13 times (87\%) whereas HV-T (7 times) and V-T (5 times) were ranked easiest to understand a combined 12 times (80\%).
    
    \item \textit{Cooperation and agreement}: We measured the user's subjective rating of the level of cooperation with the navigation assistance. Our results show that the assisted control condition type (H, V-T, V-B, HV-T, HV-B) had a significant effect on the level of cooperation, $\chi^2(4) = 11.76, p = 0.019$. A post-hoc test revealed that only V-T led to marginally significant effects over H ($p = 0.058$). We assessed the mean disagreement metric and found that the navigation assistance condition had no such effects across control conditions ($p = 0.738$). In the subjective ranking, visual trajectory-based conditions (HV-T and V-T) were ranked with most cooperation a combined 11 times (73\%), with HV-T topping the rank with 7 first place rankings (47\%). Again, the haptic (H) condition was ranked as least in cooperation 13 times (87\%).
    
    \item \textit{Sense of control}: When evaluating the participant's sense of control across control conditions, we found a significant effect of control condition on reported sense of control, $\chi^2(5) = 27.89, p = 0.000$. A post-hoc test demonstrated that MC ($p = 0.054$), V-T ($p = 0.067$) and V-B ($p = 0.081$) conditions led to marginally significant increase in perception of control compared to the H condition.
    
    \item \textit{Overall preference}: When asked to rank the control conditions in terms of overall preference, six participants (40\%) ranked HV-T condition in first place, whereas three participants each ranked V-T, HV-B and MC conditions instead (see Fig.~\ref{fig:preference}). Haptic (H) condition was ranked least preferred 11 times (73\%). 
    Table~\ref{rank-results} shows the rank order and calculated scores for overall preference as well as other subjective measures.
\end{enumerate}

\section{Discussion}\label{sec:discussion}

In this study, our proposed multi-modal shared autonomy system for assisting in social navigation was evaluated and compared with single-modality designs (haptic or visual) and manual control (no assistance) as the baseline based on metrics such as navigation performance, efficiency, interface transparency, cooperation and user preference.

\subsection{Effect on Navigation Safety and Efficiency}\label{subsec:nav-safety}

We hypothesized that multimodal assistance will lead to higher navigation performance compared to single modal assistance (haptic and visual) and our baseline condition. However, we observed that neither the presence nor the type of assistance modality had a significant effect on both the navigation performance of the telepresence robot. We compare this finding with previous studies related to assisted navigation of mobile telepresence robots. A DWA-based shared control approach by Takayama et al.~\cite{Takayama2011} reduced collisions while increasing task completion times. Macharet et al.~\cite{MacHaret2012} used potential-field-based navigation assistance to reduce collisions and completion times. The adaptive shared controller by Carlson et al.~\cite{carlson2012online} reduced completion time while participants were performing a secondary task simultaneously. These studies, however, consider assistance in static environments only (e.g. static obstacle course). Two recent studies on telepresence robot navigation assistance have considered dynamic, human-populated environments. 
Beraldo et al.~\cite{beraldo2021shared} proposed a people-aware navigation assistance approach that reduced collisions compared to manual control but did not consider personal space intrusions specifically. The study by Batmaz et al.~\cite{batmaz2020automatic} showed that assistance methods may not always improve navigation performance, especially in dynamic scenarios in interaction with people.
This paper extends current work by considering navigation assistance in dynamic, human-populated environments with a focus on shared autonomy modalities that improve system transparency, cooperation and user preference.~

A possible explanation for the lack of significant effect on navigation performance in our study is task difficulty. Kuiper et al.~\cite{Kuiper2016} found that navigation assistance systems are more effective when the task is more difficult. However, in our final questionnaire, participants reported high confidence in their ability to complete the task without assistance ($M= 5.07, SD= 1.39$). A study by Lee et al.~\cite{lee1994trust} noted that people use autonomous systems based on their trust in the system versus their self-confidence to perform the task. This suggests our work warrants further investigations in two directions. First, further experimental manipulations of the task difficulty may be needed to provide more insight into its effects on shared navigation tasks. The second direction is to investigate the effect of training and learning on operator trust and reliance on navigation assistance by manipulating practice duration.

\subsection{Effect on System Transparency and Cooperation}

Regarding system transparency and cooperation, we hypothesized that (i) multimodal assistance will result in higher rating than both single modal cases, and (ii) visual modality will result in higher rating than the haptic modality. Our results only partially support these hypotheses. In the case of intent understanding, our results revealed that the control condition type significantly impacted the level of intent understanding. Specifically, we found that visual cues delineate the intent of the autonomous agent significantly more than the haptic cues alone. This is consistent with existing work on the visual channel for intent communication~\cite{brooks2020visualization, evans2015investigating}. However, we do not see a significant difference between visual-based single modal conditions (V-T, V-B) and multi-modal conditions (HV-T, HV-B), i.e. the combination of haptic and visual cues didn't lead to higher intent understanding over visual only conditions. A possible explanation for this is that due to the dynamic nature of our task, the haptic guidance forces tended to be fast changing and sometimes caused a distraction. Some participants commented that they were unable to understand why or what the haptic condition was trying to do.

Considering force anticipation, our results show that including visual information significantly improved the force anticipation over haptic cues alone, especially when the visual cue is presented as steering magnitude bars (V-B). This is consistent with the findings by Ho et al.~\cite{ho2018increasing}. V-B provided a more consistent interpretation of the haptic forces than the suggested path visuals (V-T).

Additionally, in considering human-automation cooperation and agreement, our subjective results revealed that level of cooperation was significantly impacted by assistance condition. Participants reported least cooperation for haptic only and highest with V-T. We must reflect on these results with caution as we found no significant difference when objectively estimating the degree of agreement (via the mean disagreement measure) between haptic only and visual conditions (see Table~\ref{rank-results}). However, in the subjective ranking, we see that Borda score placed HV-T over others consistently across all factors. Nine participants commented that visual information helped them better understand the haptic forces. This is supported by existing studies on multimodal feedback~\cite{Vreugdenhil2019, van2020visual}. To our knowledge, our study is the first to evaluate the effects of different feedback guidance modalities on system transparency, cooperation and user preference in navigation around moving pedestrians.

\subsection{Effect on User Preferences}

We hypothesized that multi-modal assistance will be preferred over single modality assistance for both visual and haptic only cases. The results of our subjective ranking only partially support this hypothesis. Participants ranked both multi-modal assistance conditions, HV-T and HV-B, higher than H, with high Borda score margins in both cases, 46 and 32 points respectively (see Table~\ref{rank-results}). In fact, haptic only (H)  condition was ranked last by 73\% of participants in the study (11 times). This probably occurred because of the dynamic nature of the navigation task. Though the haptic channel enables faster response through reflexes~\cite{de2007design}, it suffers from low bandwidth limiting the amount of intent information that can be communicated to the operator~\cite{Kuiper2016}. In our results, we found that providing guidance information in visual form led to higher preferences. This is supported by existing multimodal studies~\cite{Vreugdenhil2019, van2020visual}. In the post-study interviews, nine participants reported that visual information enabled them to better understand the haptic forces.

We additionally hypothesized that multi-modal assistance will be preferred over visual only cases. Our results show only partial support for this. For visual trajectory case, although HV-T is ranked higher, it is with a weak margin (only 2 points) (see Table~\ref{rank-results}). As for visual magnitude bars case, we see the reverse case with V-B ranked higher in the overall preference by only 1 point. When asked why they prefer multi-modal assistance over visual only assistance, participants reported that though visual information helped with understanding, the haptic cues ``served as a reminder", ``pushed you a little when a person is behind you", ``moves you especially in moments of confusion on where to go".

\subsection{Effect of Visual Guidance Design}

In our study, we considered two designs for visual guidance: visual guidance trajectory and visual steering bars, as described in  Section~\ref{subsec:visual}.  According to our results, objective and subjective ratings did not differ significantly between both designs. However, in analyzing the subjective rankings, we found that trajectory conditions  (V-T \& HV-T) consistently had higher rank-score over steering bar conditions (V-B \& HV-B) (see Table~\ref{rank-results}). Specifically, visual trajectory conditions were ranked most preferred 9 times (60\%) compared to 3 times (20\%) for visual steering bar conditions. Five participants reported that the suggested path in V-T was easier to match in terms of where to go. Also, one participant remarked that the suggested path conveyed more information than the steering bars. This is consistent with the operator display design paradigms in literature~\cite{ evans2015investigating}. In~\cite{Vreugdenhil2019}, the authors suggest that a predicted trajectory expressing an autonomous agent's intent provides more information about the driving context than just an instantaneous maneuver suggestion (which reflects the haptic forces). Nevertheless, some participants reported they preferred steering bars. One reason they cited was that they found the steering bars easier to follow than the suggested path, possibly because it provided specific corrections on what control input to apply. These findings provide insights for further research into human-centered visual design options in assistive systems.

\subsection{Limitations and Future Work}

In this study, we have not evaluated the effects of trust in the shared autonomy framework as introduced in  Section~\ref{subsec:nav-safety}.  Research shows that human-robot trust influences joint task performance~\cite{hancock2011meta, lee1994trust}, with low human trust in the robot partner degrading task performance. Conversely, higher trust levels increase overall task performance, user acceptance, and reduce operator workloads~\cite{Saeidi2016}. Trust, as a dynamic attribute, is continuously calibrated based on joint interaction with the robot partner and has also been shown to depend on the robot partner's past and current performance~\cite{hancock2011meta, Saeidi2016} as well as transparency of the interface~\cite{ososky2014determinants}. We hypothesize that the lack of significant difference in navigation performance in our study may partly be due to trust as a confounding factor. As such, future investigations should measure and evaluate trust levels across modalities and explore the interaction effects with navigation performance and team cooperation.


Additionally, our small sample size may have limited our ability to identify significant differences and strong effects in our study due to our large number of comparisons explored across control conditions.  Furthermore, our sample population comprised primarily young adults. We acknowledge that the lack of age-related diversity in the study may limit the generalizability of our findings.  Even so, our study provides a good foundation for future research in understanding the effects of multimodal assistance systems in remote teleoperation in dynamic, human-populated spaces.


Furthermore, we acknowledge the discrepancy between simulated and real-world experiments. Simulated experiments have been used extensively in the literature~\cite{Zhang2020, Abbink2018, Saeidi2016} to study human-machine interactions in driving tasks as they allow for more control of experimental conditions, relax human safety concerns, and still provide a method of investigating human-machine interactions effectively. However, we recognize that a real world navigation study is needed as next steps of this work to further validate our hypotheses. Therefore, future work will include addressing real-robot implementation such as robot dynamic constraints, perception and SLAM using robot noisy onboard sensors as well as real-world testing in crowded human spaces (e.g., cafeteria).

\section{Conclusion}\label{sec7}

In this paper, we presented a multi-modal shared autonomy approach for navigation assistance in dynamic, human-populated environments. This approach included both active haptic guidance and passive visual feedback by using two distinct visualizations of the safe control input. We conducted a user study (n=15) to evaluate how our proposed multimodal approach affects navigation performance, interface transparency, and user preference compared to single modality (haptic or visual alone) and no assistance cases. Our results revealed that more operators preferred multi-modal assistance (especially when visualization is in trajectory form) over both visual or haptic only in the shared navigation task. While we did not find any significant differences in navigation performance, we did find that visual cues significantly increased participants' understanding of intent and level of cooperation over haptic guidance. In this study, we have made important advances towards our understanding of how to design navigation assistance systems within the complex context of socially-aware navigation in cluttered environments. We conclude that further research is needed to validate these findings in a real-world context.



\ifCLASSOPTIONcaptionsoff
  \newpage
\fi



\bibliographystyle{IEEEtran}

\bibliography{references}

\begin{thebibliography}{10}
\providecommand{\url}[1]{#1}
\csname url@samestyle\endcsname
\providecommand{\newblock}{\relax}
\providecommand{\bibinfo}[2]{#2}
\providecommand{\BIBentrySTDinterwordspacing}{\spaceskip=0pt\relax}
\providecommand{\BIBentryALTinterwordstretchfactor}{4}
\providecommand{\BIBentryALTinterwordspacing}{\spaceskip=\fontdimen2\font plus
\BIBentryALTinterwordstretchfactor\fontdimen3\font minus
  \fontdimen4\font\relax}
\providecommand{\BIBforeignlanguage}[2]{{%
\expandafter\ifx\csname l@#1\endcsname\relax
\typeout{** WARNING: IEEEtran.bst: No hyphenation pattern has been}%
\typeout{** loaded for the language `#1'. Using the pattern for}%
\typeout{** the default language instead.}%
\else
\language=\csname l@#1\endcsname
\fi
#2}}
\providecommand{\BIBdecl}{\relax}
\BIBdecl

\bibitem{Neustaedter2018}
C.~Neustaedter, S.~Singhal, R.~Pan, Y.~Heshmat, A.~Forghani, and J.~Tang,
  ``{From Being There to Watching: Shared and Dedicated Telepresence Robot
  Usage at Academic Conferences},'' \emph{ACM Trans. Comput.-Hum. Interact},
  vol.~25, 2018.

\bibitem{Khojasteh2019}
\BIBentryALTinterwordspacing
N.~Khojasteh, C.~Liu, and S.~R. Fussell, ``{Understanding undergraduate
  students' experiences of telepresence robots on campus},'' \emph{Proceedings
  of the ACM Conference on Computer Supported Cooperative Work, CSCW}, pp.
  241--246, 2019. [Online]. Available:
  \url{https://doi.org/10.1145/3311957.3359450.}
\BIBentrySTDinterwordspacing

\bibitem{mimnaugh2021analysis}
K.~J. Mimnaugh, M.~Suomalainen, I.~Becerra, E.~Lozano, R.~Murrieta-Cid, and
  S.~M. LaValle, ``Analysis of user preferences for robot motions in immersive
  telepresence,'' \emph{arXiv preprint arXiv:2103.03496}, 2021.

\bibitem{Kruse2013}
T.~Kruse, A.~K. Pandey, R.~Alami, and A.~Kirsch, ``{Human-aware robot
  navigation: A survey},'' \emph{Robotics and Autonomous Systems}, vol.~61,
  no.~12, pp. 1726--1743, dec 2013.

\bibitem{mavrogiannis2021core}
C.~Mavrogiannis, F.~Baldini, and et~al, ``Core challenges of social robot
  navigation: A survey,'' \emph{arXiv preprint arXiv:2103.05668}, 2021.

\bibitem{Takayama2011}
L.~Takayama, E.~Marder-Eppstein, H.~Harris, and J.~M. Beer, ``{Assisted driving
  of a mobile remote presence system: System design and controlled user
  evaluation},'' in \emph{Proceedings - IEEE International Conference on
  Robotics and Automation}, 2011, pp. 1883--1889.

\bibitem{Abbink2018}
D.~A. Abbink, T.~Carlson, M.~Mulder, J.~C. {De Winter}, F.~Aminravan, T.~L.
  Gibo, and E.~R. Boer, ``{A topology of shared control systems-finding common
  ground in diversity},'' \emph{IEEE Transactions on Human-Machine Systems},
  vol.~48, no.~5, pp. 509--525, oct 2018.

\bibitem{Petermeijer2015}
S.~M. Petermeijer, D.~A. Abbink, M.~Mulder, and J.~C. {De Winter}, ``{The
  Effect of Haptic Support Systems on Driver Performance: A Literature
  Survey},'' pp. 467--479, oct 2015.

\bibitem{Itoh2016}
\BIBentryALTinterwordspacing
M.~Itoh, F.~Flemisch, and D.~Abbink, ``{A hierarchical framework to analyze
  shared control conflicts between human and machine},''
  \emph{IFAC-PapersOnLine}, vol.~49, no.~19, pp. 96--101, 2016. [Online].
  Available: \url{www.sciencedirect.com}
\BIBentrySTDinterwordspacing

\bibitem{Muelling2018}
\BIBentryALTinterwordspacing
K.~Muelling, C.~Takahashi, S.~Nikolaidis, V.~Alonso, and P.~{De La Puente},
  ``{System Transparency in Shared Autonomy: A Mini Review},'' \emph{Frontiers
  in Neurorobotics | www.frontiersin.org}, vol.~12, p.~83, 2018. [Online].
  Available: \url{www.frontiersin.org}
\BIBentrySTDinterwordspacing

\bibitem{de2007design}
S.~De~Stigter, M.~Mulder, and M.~Van~Paassen, ``Design and evaluation of a
  haptic flight director,'' \emph{Journal of guidance, control, and dynamics},
  vol.~30, no.~1, pp. 35--46, 2007.

\bibitem{Kuiper2016}
R.~J. Kuiper, D.~J. Heck, I.~A. Kuling, and D.~A. Abbink, ``{Evaluation of
  Haptic and Visual Cues for Repulsive or Attractive Guidance in Nonholonomic
  Steering Tasks},'' \emph{IEEE Transactions on Human-Machine Systems},
  vol.~46, no.~5, pp. 672--683, oct 2016.

\bibitem{Zolotas2019}
M.~Zolotas and Y.~Demiris, ``{Towards Explainable Shared Control using
  Augmented Reality},'' in \emph{IEEE International Conference on Intelligent
  Robots and Systems}.\hskip 1em plus 0.5em minus 0.4em\relax Institute of
  Electrical and Electronics Engineers Inc., nov 2019, pp. 3020--3026.

\bibitem{Brooks2020}
\BIBentryALTinterwordspacing
C.~Brooks and D.~Szafir, ``{Visualization of Intended Assistance for Acceptance
  of Shared Control},'' in \emph{IROS 2020}, 2020. [Online]. Available:
  \url{http://arxiv.org/abs/2008.10759}
\BIBentrySTDinterwordspacing

\bibitem{Vreugdenhil2019}
W.~Vreugdenhil, S.~Barendswaard, D.~A. Abbink, C.~Borst, and S.~M. Petermeijer,
  ``{Complementing Haptic Shared Control with Visual Feedback for Obstacle
  Avoidance},'' in \emph{IFAC-PapersOnLine}, vol.~52, no.~19.\hskip 1em plus
  0.5em minus 0.4em\relax Elsevier B.V., jan 2019, pp. 371--376.

\bibitem{van2020visual}
A.~van~den Berg, ``Visual feedback for haptic assisted teleoperation of an
  industrial robot: With dross removal as a use case,'' 2020.

\bibitem{hall1966hidden}
E.~T. Hall, \emph{The hidden dimension}.\hskip 1em plus 0.5em minus 0.4em\relax
  Garden City, NY: Doubleday, 1966, vol. 609.

\bibitem{socnavassist21}
K.~C. Mbanisi, M.~Gennert, and Z.~Li, ``Socnavassist: A haptic shared autonomy
  framework for social navigation assistance of mobile telepresence robots,''
  in \emph{2021 IEEE 2nd International Conference on Human-Machine Systems
  (ICHMS)}, 2021, pp. 1--3.

\bibitem{Beer2014}
J.~M. Beer, A.~D. Fisk, and W.~A. Rogers, ``{Toward a Framework for Levels of
  Robot Autonomy in Human-Robot Interaction},'' \emph{Journal of Human-Robot
  Interaction}, vol.~3, no.~2, pp. 74--99, 2014.

\bibitem{rios2011understanding}
J.~Rios-Martinez, A.~Spalanzani, and C.~Laugier, ``Understanding human
  interaction for probabilistic autonomous navigation using risk-rrt
  approach,'' in \emph{2011 IEEE/RSJ International Conference on Intelligent
  Robots and Systems}.\hskip 1em plus 0.5em minus 0.4em\relax IEEE, 2011, pp.
  2014--2019.

\bibitem{helbing1995social}
D.~Helbing and P.~Molnar, ``Social force model for pedestrian dynamics,''
  \emph{Physical review E}, vol.~51, no.~5, p. 4282, 1995.

\bibitem{VanBerg2008}
\BIBentryALTinterwordspacing
J.~D. {Van Berg}, M.~Lin, and D.~Manocha, ``{Reciprocal velocity obstacles for
  real-time multi-agent navigation},'' in \emph{Proceedings - IEEE
  International Conference on Robotics and Automation}, 2008, pp. 1928--1935.
  [Online]. Available: \url{http://gamma.cs.unc.edu/RVO.}
\BIBentrySTDinterwordspacing

\bibitem{Chen2017}
Y.~F. Chen, M.~Everett, M.~Liu, and J.~P. How, ``{Socially aware motion
  planning with deep reinforcement learning},'' in \emph{IEEE International
  Conference on Intelligent Robots and Systems}, vol. 2017-Septe.\hskip 1em
  plus 0.5em minus 0.4em\relax Institute of Electrical and Electronics
  Engineers Inc., dec 2017, pp. 1343--1350.

\bibitem{Narayanan2016}
V.~K. Narayanan, A.~Spalanzani, and M.~Babel, ``{A semi-autonomous framework
  for human-aware and user intention driven wheelchair mobility assistance},''
  in \emph{IEEE International Conference on Intelligent Robots and Systems},
  vol. 2016-Novem.\hskip 1em plus 0.5em minus 0.4em\relax Institute of
  Electrical and Electronics Engineers Inc., nov 2016, pp. 4700--4707.

\bibitem{Kretzschmar2016}
H.~Kretzschmar, M.~Spies, C.~Sprunk, and W.~Burgard, ``{Socially compliant
  mobile robot navigation via inverse reinforcement learning},'' \emph{The
  International Journal of Robotics Research}, vol.~35, no.~11, pp. 1289--1307,
  2016.

\bibitem{hoc2009cooperation}
J.-M. Hoc, M.~S. Young, and J.-M. Blosseville, ``Cooperation between drivers
  and automation: implications for safety,'' \emph{Theoretical Issues in
  Ergonomics Science}, vol.~10, no.~2, pp. 135--160, 2009.

\bibitem{losey2018review}
D.~P. Losey, C.~G. McDonald, E.~Battaglia, and M.~K. O'Malley, ``A review of
  intent detection, arbitration, and communication aspects of shared control
  for physical human--robot interaction,'' \emph{Applied Mechanics Reviews},
  vol.~70, no.~1, 2018.

\bibitem{Abbink}
D.~A. Abbink, M.~Mulder, and E.~R. Boer, ``{Haptic shared control: smoothly
  shifting control authority?}''

\bibitem{Zhang2020}
D.~Zhang, G.~Yang, and R.~P. Khurshid, ``{Haptic Teleoperation of UAVs through
  Control Barrier Functions},'' \emph{IEEE Transactions on Haptics}, vol.~13,
  no.~1, pp. 109--115, jan 2020.

\bibitem{ho2018increasing}
V.~Ho, C.~Borst, M.~M. van Paassen, and M.~Mulder, ``Increasing acceptance of
  haptic feedback in uav teleoperation by visualizing force fields,'' in
  \emph{2018 IEEE International Conference on Systems, Man, and Cybernetics
  (SMC)}.\hskip 1em plus 0.5em minus 0.4em\relax IEEE, 2018, pp. 3027--3032.

\bibitem{Truong2017}
X.~T. Truong and T.~D. Ngo, ``{Toward Socially Aware Robot Navigation in
  Dynamic and Crowded Environments: A Proactive Social Motion Model},''
  \emph{IEEE Transactions on Automation Science and Engineering}, vol.~14,
  no.~4, pp. 1743--1760, oct 2017.

\bibitem{jain2019probabilistic}
S.~Jain and B.~Argall, ``Probabilistic human intent recognition for shared
  autonomy in assistive robotics,'' \emph{ACM Transactions on Human-Robot
  Interaction (THRI)}, vol.~9, no.~1, pp. 1--23, 2019.

\bibitem{linder2014multi}
T.~Linder and K.~O. Arras, ``Multi-model hypothesis tracking of groups of
  people in rgb-d data,'' in \emph{17th International conference on information
  fusion (FUSION)}.\hskip 1em plus 0.5em minus 0.4em\relax IEEE, 2014, pp.
  1--7.

\bibitem{brooks2020visualization}
C.~Brooks and D.~Szafir, ``Visualization of intended assistance for acceptance
  of shared control,'' in \emph{2020 IEEE/RSJ International Conference on
  Intelligent Robots and Systems (IROS)}.\hskip 1em plus 0.5em minus
  0.4em\relax IEEE, 2020, pp. 11\,425--11\,430.

\bibitem{evans2015investigating}
A.~W. Evans~III, S.~G. Hill, and R.~Pomranky, ``Investigating the usefulness of
  soldier aids for autonomous unmanned ground vehicles, part 2,'' ARMY RESEARCH
  LAB ABERDEEN PROVING GROUND MD HUMAN RESEARCH AND ENGINEERING~…, Tech.
  Rep., 2015.

\bibitem{wise2016fetch}
M.~Wise, M.~Ferguson, D.~King, E.~Diehr, and D.~Dymesich, ``Fetch and freight:
  Standard platforms for service robot applications,'' in \emph{Workshop on
  autonomous mobile service robots}, 2016.

\bibitem{fukao2000adaptive}
T.~Fukao, H.~Nakagawa, and N.~Adachi, ``Adaptive tracking control of a
  nonholonomic mobile robot,'' \emph{IEEE transactions on Robotics and
  Automation}, vol.~16, no.~5, pp. 609--615, 2000.

\bibitem{rios2015proxemics}
J.~Rios-Martinez, A.~Spalanzani, and C.~Laugier, ``From proxemics theory to
  socially-aware navigation: A survey,'' \emph{International Journal of Social
  Robotics}, vol.~7, no.~2, pp. 137--153, 2015.

\bibitem{rsoft}
\BIBentryALTinterwordspacing
{R Core Team}, \emph{R: A Language and Environment for Statistical Computing},
  R Foundation for Statistical Computing, Vienna, Austria, 2021. [Online].
  Available: \url{https://www.R-project.org/}
\BIBentrySTDinterwordspacing

\bibitem{MacHaret2012}
D.~G. MacHaret and D.~A. Florencio, ``{A collaborative control system for
  telepresence robots},'' in \emph{IEEE International Conference on Intelligent
  Robots and Systems}, 2012, pp. 5105--5111.

\bibitem{carlson2012online}
T.~Carlson, R.~Leeb, R.~Chavarriaga, and J.~d.~R. Mill{\'a}n, ``Online
  modulation of the level of assistance in shared control systems,'' in
  \emph{2012 IEEE International Conference on Systems, Man, and Cybernetics
  (SMC)}.\hskip 1em plus 0.5em minus 0.4em\relax IEEE, 2012, pp. 3339--3344.

\bibitem{beraldo2021shared}
G.~Beraldo, K.~Koide, A.~Cesta, S.~Hoshino, J.~Miura, M.~Salv{\`a}, and
  E.~Menegatti, ``Shared autonomy for telepresence robots based on people-aware
  navigation,'' in \emph{International Conference on Intelligent Autonomous
  Systems}.\hskip 1em plus 0.5em minus 0.4em\relax Springer, 2021, pp.
  109--122.

\bibitem{batmaz2020automatic}
A.~U. Batmaz, J.~Maiero, E.~Kruijff, B.~E. Riecke, C.~Neustaedter, and
  W.~Stuerzlinger, ``How automatic speed control based on distance affects user
  behaviours in telepresence robot navigation within dense conference-like
  environments,'' \emph{Plos one}, vol.~15, no.~11, p. e0242078, 2020.

\bibitem{lee1994trust}
J.~D. Lee and N.~Moray, ``Trust, self-confidence, and operators' adaptation to
  automation,'' \emph{International journal of human-computer studies},
  vol.~40, no.~1, pp. 153--184, 1994.

\bibitem{hancock2011meta}
P.~A. Hancock, D.~R. Billings, K.~E. Schaefer, J.~Y. Chen, E.~J. De~Visser, and
  R.~Parasuraman, ``A meta-analysis of factors affecting trust in human-robot
  interaction,'' \emph{Human factors}, vol.~53, no.~5, pp. 517--527, 2011.

\bibitem{Saeidi2016}
H.~Saeidi, F.~McLane, B.~Sadrfaidpour, E.~Sand, S.~Fu, J.~Rodriguez, J.~R.
  Wagner, and Y.~Wang, ``{Trust-based mixed-initiative teleoperation of mobile
  robots},'' in \emph{Proceedings of the American Control Conference}, vol.
  2016-July.\hskip 1em plus 0.5em minus 0.4em\relax Institute of Electrical and
  Electronics Engineers Inc., jul 2016, pp. 6177--6182.

\bibitem{ososky2014determinants}
S.~Ososky, T.~Sanders, F.~Jentsch, P.~Hancock, and J.~Y. Chen, ``Determinants
  of system transparency and its influence on trust in and reliance on unmanned
  robotic systems,'' in \emph{Unmanned Systems Technology XVI}, vol.
  9084.\hskip 1em plus 0.5em minus 0.4em\relax International Society for Optics
  and Photonics, 2014, p. 90840E.

\end{thebibliography}
\end{document}